\documentclass[10pt,twocolumn,letterpaper]{article}

\usepackage{cvpr}
\usepackage{times}
\usepackage{epsfig}
\usepackage{graphicx}
\usepackage{amsmath}
\usepackage{amssymb}

\usepackage{booktabs}       

\usepackage{amsmath}
\usepackage{mathtools}
\usepackage{pgfplots}
\usepackage{filecontents}
\usepackage{multirow}
\usepackage{xcolor}
\usepackage[percent]{overpic}
\usepackage{placeins}
\usepackage{afterpage}
\usepackage[shortlabels]{enumitem}

\setlist[enumerate]{itemsep=1pt}
\makeatletter
\renewcommand\paragraph{\@startsection{paragraph}{4}{\z@}%
                                      {.5em \@plus.2ex \@minus.2ex}%
                                      {-1em}%
                                      {\normalfont\normalsize\bfseries}}

\makeatother
\definecolor{blue}{RGB}{4, 93, 143}
\definecolor{orange}{RGB}{237, 125, 49}

\pgfplotsset{compat=1.14}
\pgfplotsset{every axis plot/.append style={ultra thick}, legend style={draw=none},}
\pgfplotsset{axis x line=bottom, axis y line=left}

\usepackage[pagebackref=true,breaklinks=true,letterpaper=true,colorlinks,bookmarks=false]{hyperref}

\def\cvprPaperID{5463} 

\def\R#1{{\mathbb{R}^{#1}}}
\def\transpose{^\top}
\def\imagemap{\mathbf{f}}
\def\voxelmap{\mathbf{g}}
\def\orthomap{\mathbf{h}}
\def\integralimg{\mathbf{F}}
\def\confmap{S}
\def\res{r}
\def\posoffset{\boldsymbol{\Delta_{pos}}}
\def\dimoffset{\boldsymbol{\Delta_{dim}}}
\def\angoffset{\boldsymbol{\Delta_{ang}}}

\def\figref#1{Figure~\ref{fig:#1}}
\def\secref#1{Section~\ref{sec:#1}}
\def\eqnref#1{Equation~\ref{eqn:#1}}
\def\tabref#1{Table~\ref{tab:#1}}

\cvprfinalcopy

\title{Orthographic Feature Transform for Monocular 3D Object Detection}

\author{
  Thomas Roddick \qquad Alex Kendall \qquad Roberto Cipolla\\
 University of Cambridge \\
 \small\texttt{\{tr346, agk34, rc10001\}@cam.ac.uk} \\
}

\begin{document}
\maketitle

\global\csname @topnum\endcsname 0
\global\csname @botnum\endcsname 0

\begin{figure}[t]
  \centering
  \includegraphics[width=\linewidth]{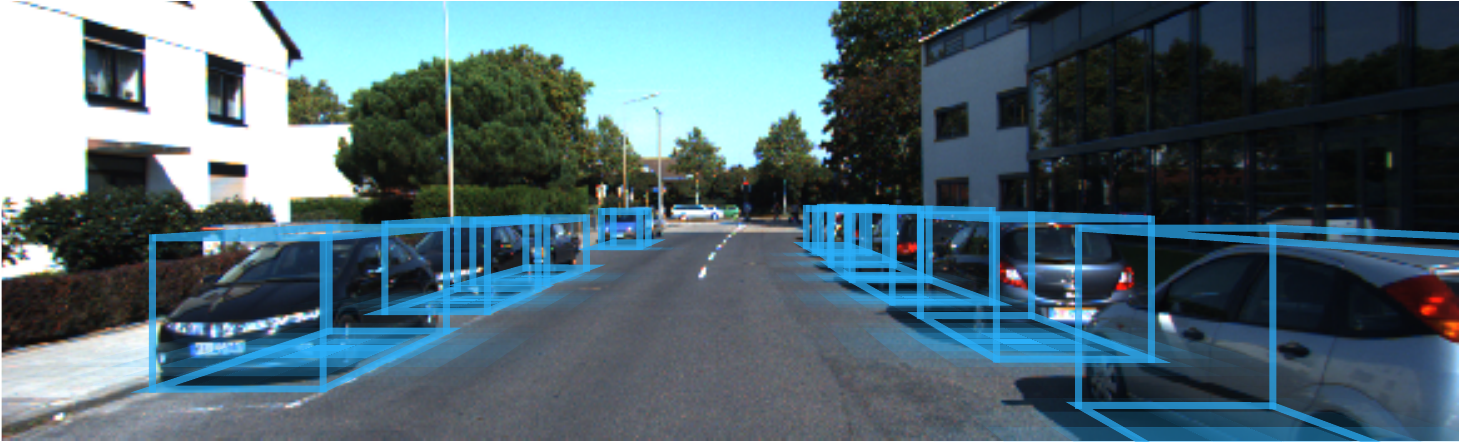}
  \caption{3D bounding box detection from monocular images. The proposed system maps image-based features to an orthographic birds-eye-view and predicts confidence maps and bounding box offsets in this space. These outputs are then decoded via non-maximum suppression to yield discrete bounding box predictions.}
  \label{fig:arch}
\end{figure}

\begin{abstract}
3D object detection from monocular images has proven to be an enormously challenging task, with the performance of leading systems not yet achieving even 10\% of that of LiDAR-based counterparts. One explanation for this performance gap is that existing systems are entirely at the mercy of the perspective image-based representation, in which the appearance and scale of objects varies drastically with depth and meaningful distances are difficult to infer. In this work we argue that the ability to reason about the world in 3D is an essential element of the 3D object detection task. To this end, we introduce the orthographic feature transform, which enables us to escape the image domain by mapping image-based features into an orthographic 3D space. This allows us to reason holistically about the spatial configuration of the scene in a domain where scale is consistent and distances between objects are meaningful. We apply this transformation as part of an end-to-end deep learning architecture and achieve state-of-the-art performance on the KITTI 3D object benchmark.\footnote{We will release full source code and pretrained models upon acceptance of this manuscript for publication.}

\end{abstract}


\section{Introduction}
\label{sec:intro}
The success of any autonomous agent is contingent on its ability to detect and localize the objects in its surrounding environment. Prediction, avoidance and path planning all depend on robust estimates of the 3D positions and dimensions of other entities in the scene. This has led to 3D bounding box detection emerging as an important problem in computer vision and robotics, particularly in the context of autonomous driving. To date the 3D object detection literature has been dominated by approaches which make use of rich LiDAR point clouds~\cite{yu2017vehicle, wirges2018object, ku2017joint, qi2017frustum, chen2017multi, du2018general, minemura2018lmnet, beltran2018birdnet}, while the performance of image-only methods, which lack the absolute depth information of LiDAR, lags significantly behind. Given the high cost of existing LiDAR units, the sparsity of LiDAR point clouds at long ranges, and the need for sensor redundancy, accurate 3D object detection from monocular images remains an important research objective. To this end, we present a novel 3D object detection algorithm which takes a single monocular RGB image as input and produces high quality 3D bounding boxes, achieving state-of-the-art performance among monocular methods on the challenging KITTI benchmark~\cite{geiger2012kitti}.

Images are, in many senses, an extremely challenging modality. Perspective projection implies that the scale of a single object varies considerably with distance from the camera; its appearance can change drastically depending on the viewpoint; and distances in the 3D world cannot be inferred directly. These factors present enormous challenges to a monocular 3D object detection system. A far more innocuous representation is the orthographic birds-eye-view map commonly employed in many LiDAR-based methods~\cite{yu2017vehicle, wirges2018object, beltran2018birdnet}. Under this representation, scale is homogeneous; appearance is largely viewpoint-independent; and distances between objects are meaningful. Our key insight therefore is that as much reasoning as possible should be performed in this orthographic space rather than directly on the pixel-based image domain. This insight proves essential to the success of our proposed system.

It is unclear, however, how such a representation could be constructed from a monocular image alone. We therefore introduce the \emph{orthographic feature transform} (OFT): a differentiable transformation which maps a set of features extracted from a perspective RGB image to an orthographic birds-eye-view feature map. Crucially, we do not rely on any explicit notion of depth: rather our system builds up an internal representation which is able to determine which features from the image are relevant to each location on the birds-eye-view. We apply a deep convolutional neural network, the \emph{topdown} network, in order to reason locally about the 3D configuration of the scene.

The main contributions of our work are as follows:
\begin{enumerate}
    \item We introduce the orthographic feature transform (OFT) which maps perspective image-based features into an orthographic birds-eye-view, implemented efficiently using integral images for fast average pooling.
    \item We describe a deep learning architecture for predicting 3D bounding boxes from monocular RGB images.
    \item We highlight the importance of reasoning in 3D for the object detection task.
\end{enumerate}
The system is evaluated on the challenging KITTI 3D object benchmark and achieves state-of-the-art results among monocular approaches. 

\section{Related Work}
\label{sec:related}

\paragraph{2D object detection}
Detecting 2D bounding boxes in images is a widely studied problem and recent approaches are able to excel even on the most formidable datasets~\cite{russakovsky2015imagenet, everingham2010pascal, lin2014microsoft}. Existing methods may broadly be divided into two main categories: \emph{single stage} detectors such as YOLO~\cite{redmon2018yolov3}, SSD~\cite{liu2016ssd} and RetinaNet~\cite{lin2018focal} which predict object bounding boxes directly and \emph{two-stage} detectors such as Faster R-CNN~\cite{ren2015faster} and FPN~\cite{lin2017feature} which add an intermediate region proposal stage. To date the vast majority of 3D object detection methods have adopted the latter philosophy, in part due to the difficulty in mapping from fixed-sized regions in 3D space to variable-sized regions in the image space. We overcome this limitation via our OFT transform, allowing us to take advantage of the purported speed and accuracy benefits~\cite{lin2018focal} of a single-stage architecture.

\paragraph{3D object detection from LiDAR}
3D object detection is of considerable importance to autonomous driving, and a large number of LiDAR-based methods have been proposed which have enjoyed considerable success. Most variation arises from how the LiDAR point clouds are encoded. The Frustrum-PointNet of Qi~\etal~\cite{qi2017frustum} and the work of Du~\etal~\cite{du2018general} operate directly on the point clouds themselves, considering a subset of points which lie within a frustrum defined by a 2D bounding box on the image. Minemura~\etal~\cite{minemura2018lmnet} and Li~\etal~\cite{li2016vehicle} instead project the point cloud onto the image plane and apply Faster-RCNN-style architectures to the resulting RGB-D images. Other methods, such as TopNet~\cite{wirges2018object}, BirdNet~\cite{beltran2018birdnet} and Yu~\etal~\cite{yu2017vehicle}, discretize the point cloud into some birds-eye-view (BEV) representation which encodes features such as returned intensity or average height of points above the ground plane. This representation turns out to be extremely attractive since it does not exhibit any of the perspective artifacts introduced in RGB-D images for example, and a major focus of our work is therefore to develop an implicit image-only analogue to these birds-eye-view maps. A further interesting line of research is sensor fusion methods such as AVOD~\cite{ku2017joint} and MV3D~\cite{chen2017multi} which make use of 3D object proposals on the ground plane to aggregate both image-based and birds-eye-view features: an operation which is closely related to our orthographic feature transform.

\paragraph{3D object detection from images}
Obtaining 3D bounding boxes from images, meanwhile, is a much more challenging problem on account of the absence of absolute depth information. Many approaches start from 2D bounding boxes extracted using standard  detectors described above, upon which they either directly regress 3D pose parameters for each region~\cite{kehl2017ssd,poirson2016fast,novak2017vehicle,mousavian20173d} or fit 3D templates to the image~\cite{chabot2017deep, xiang2015data, xiang2017subcategory, zeeshan2014cars}. Perhaps most closely related to our work is Mono3D~\cite{chen2016monocular} which densely spans the 3D space with 3D bounding box proposals and then scores each using a variety of image-based features. Other works which explore the idea of dense 3D proposals in the world space are 3DOP~\cite{chen20153d} and Pham and Jeon~\cite{pham2017robust}, which rely on explicit estimates of depth using stereo geometry. A major limitation of all the above works is that each region proposal or bounding box is treated independently, precluding any joint reasoning about the 3D configuration of the scene. Our method performs a similar feature aggregation step to~\cite{chen2016monocular}, but applies a secondary convolutional network to the resulting proposals whilst retaining their spatial configuration. 

\paragraph{Integral images}
Integral images have been fundamentally associated with object detection ever since their introduction in the seminal work of Viola and Jones~\cite{viola2001rapid}. They have formed an important component in many contemporary 3D object detection approaches including AVOD~\cite{ku2017joint}, MV3D~\cite{chen2017multi}, Mono3D~\cite{chen2016monocular} and 3DOP~\cite{chen20153d}. In all of these cases however, integral images do not backpropagate gradients or form part of a fully end-to-end deep learning architecture. To our knowledge, the only prior work to do so is that of Kasagi~\etal~\cite{kasagi2017fast}, which combines a convolutional layer and an average pooling layer to reduce computational cost.

\FloatBarrier
\section{3D Object Detection Architecture}

In this section we describe our full approach for extracting 3D bounding boxes from monocular images. An overview of the system is illustrated in Figure~\ref{fig:arch}. The algorithm comprises five main components:
\begin{enumerate}
\item A front-end ResNet~\cite{he2016deep} feature extractor which extracts multi-scale feature maps from the input image.
\item A orthographic feature transform which transforms the image-based feature maps at each scale into an orthographic birds-eye-view representation.
\item A \emph{topdown} network, consisting of a series of ResNet residual units, which processes the birds-eye-view feature maps in a manner which is invariant to the perspective effects observed in the image.
\item A set of output heads which generate, for each object class and each location on the ground plane, a confidence score, position offset, dimension offset and a orientation vector.
\item A non-maximum suppression and decoding stage, which identifies peaks in the confidence maps and generates discrete bounding box predictions.
\end{enumerate}

The remainder of this section will describe each of these components in detail.

\subsection{Feature extraction}
The first element of our architecture is a convolutional feature extractor which generates a hierarchy of multi-scale 2D feature maps from the raw input image. These features encode information about low-level structures in the image, which form the basic components used by the topdown network to construct an implicit 3D representation of the scene. The front-end network is also responsible for inferring depth information based on the size of image features since subsequent stages of the architecture aim to eliminate variance to scale. 




\subsection{Orthographic feature transform}
\label{sec:oft}

In order to reason about the 3D world in the absence of perspective effects, we must first apply a mapping from feature maps extracted in the image space to orthographic feature maps in the world space, which we term the Orthographic Feature Transform (OFT). 

\begin{figure}[t!]
  \centering
  \includegraphics[width=\linewidth]{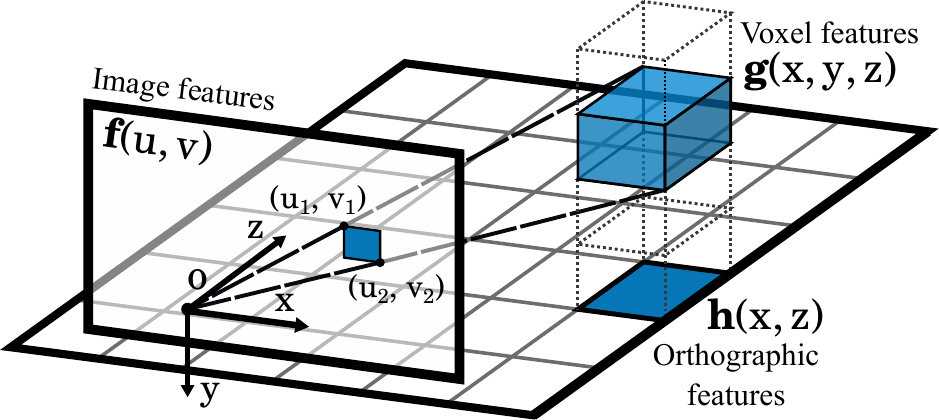}
  \caption{Orthographic Feature Transform (OFT). Voxel-based features $\voxelmap(x, y, z)$ are generated by accumulating image-based features $\imagemap(u, v)$ over the projected voxel area. The voxel features are then collapsed along the vertical dimension to yield orthographic ground plane features $\orthomap(x, z)$.}
  \label{fig:oft}
\end{figure}

The objective of the OFT is to populate the 3D \emph{voxel} feature map $\voxelmap(x, y, z)\in\R{n}$ with relevant $n$-dimensional features from the \emph{image}-based feature map $\imagemap(u, v)\in\R{n}$ extracted by the front-end feature extractor. The voxel map is defined over a uniformly spaced 3D lattice $\mathcal{G}$ which is fixed to the ground plane a distance $y_0$ below the camera and has dimensions $W$, $H$, $D$ and a voxel size of $\res$. For a given voxel grid location $(x, y, z)\in\mathcal{G}$, we obtain the voxel feature $\voxelmap(x, y, z)$ by accumulating features over the area of the image feature map $\imagemap$ which corresponds to the voxel's 2D projection. In general each voxel, which is a cube of size $\res$, will project to hexagonal region in the image plane. We approximate this by a rectangular bounding box with top-left and bottom-right corners $(u_1, v_1)$ and $(u_2, v_2)$ which are given by

\begin{equation}
\begin{split}
    u_1 &= f\frac{x-0.5\res}{z+0.5\frac{x}{|x|}\res} + c_u, \\
    u_2 &= f\frac{x+0.5\res}{z-0.5\frac{x}{|x|}\res} + c_u,
\end{split}
\quad
\begin{split}
    v_1 &= f\frac{y-0.5\res}{z+0.5\frac{y}{|y|}\res} + c_v, \\
    v_2 &= f\frac{y+0.5\res}{z-0.5\frac{y}{|y|}\res} + c_v
\end{split}
\label{eqn:corners}
\end{equation}
where $f$ is the camera focal length and $(c_u, c_v)$ the principle point. 

We can then assign a feature to the appropriate location in the voxel feature map $\voxelmap$ by average pooling over the projected voxel's bounding box in the image feature map $\imagemap$:
\begin{equation}
    \voxelmap(x, y, z) = \frac{1}{(u_2 - u_1)(v_2-v_1)}\sum_{u=u_1}^{u_2}\sum_{v=v_1}^{v_2} \imagemap(u, v)
\end{equation}

The resulting voxel feature map $\voxelmap$ already provides a representation of the scene which is free from the effects of perspective projection. However deep neural networks which operate on large voxel grids are typically extremely memory intensive. Given that we are predominantly interested in applications such as autonomous driving where most objects are fixed to the 2D ground plane, we can make the problem more tractable by collapsing the 3D voxel feature map down to a third, two-dimensional representation which we term the \emph{orthographic} feature map $\orthomap(x, z)$. The orthographic feature map is obtained by summing voxel features along the vertical axis after multiplication with a set of learned weight matrices $W(y)\in\R{n\times n}$:
\begin{equation}
    \orthomap(x, z) = \sum_{y=y_0}^{y_0+H}W(y)\voxelmap(x, y, z)
\end{equation}
Transforming to an intermediate voxel representation before collapsing to the final orthographic feature map has the advantage that the information about the vertical configuration of the scene is retained. This turns out to be essential for downstream tasks such as estimating the height and vertical position of object bounding boxes. 

\begin{figure*}[t]
  \centering
  \includegraphics[width=\textwidth]{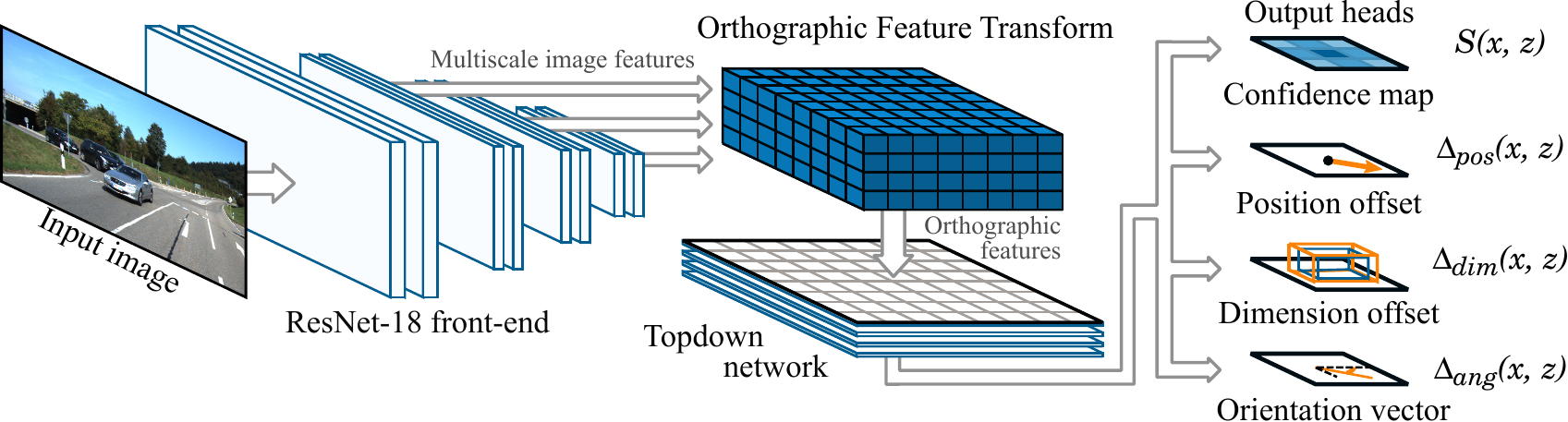}
  \caption{Architecture overview. A front-end ResNet feature extractor generates image-based features, which are mapped to an orthographic representation via our proposed orthographic feature transform. The topdown network processes these features in the birds-eye-view space and at each location on the ground plane predicts a confidence score $\confmap$, a position offset $\posoffset$, a dimension offset $\dimoffset$ and an angle vector $\angoffset$. }
  \label{fig:arch}
\end{figure*}

\subsubsection{Fast average pooling with integral images}
A major challenge with the above approach is the need to aggregate features over a very large number of regions. A typical voxel grid setting generates around 150k bounding boxes, which far exceeds the $\sim$2k regions of interest used by the Faster R-CNN~\cite{ren2015faster} architecture, for example. To facilitate pooling over such a large number of regions, we make use of a fast average pooling operation based on integral images~\cite{viola2001rapid}. An integral image, or in this case integral feature map, $\integralimg$, is constructed from an input feature map $\imagemap$ using the recursive relation
\begin{equation}
\integralimg(u, v) = \imagemap(u, v) + \integralimg(u-1, v) + \integralimg(u, v-1) - \integralimg(u-1, v-1).
\end{equation}
Given the integral feature map $\integralimg$, the output feature $\voxelmap(x, y, z)$ corresponding to the region defined by bounding box coordinates $(u_1, v_1)$ and $(u_2, v_2)$ (see \eqnref{corners}), is given by
\begin{equation}
\voxelmap(x, y, z) = \frac{\integralimg(u_1, v_1) + \integralimg(u_2, v_2) - \integralimg(u_1, v_2) - \integralimg(u_2, v_1)}{(u_2-u_1)(v_2-v_1)}
\end{equation}
The complexity of this pooling operation is independent of the size of the individual regions, which makes it highly appropriate for our application where the size and shape of the regions varies considerably depending on whether the voxel is close to or far from the camera. It is also fully differentiable in terms of the original feature map $\imagemap$ and so can be used as part of an end-to-end deep learning framework.

\subsection{Topdown network}
\label{sec:ortho}
A core contribution of this work is to emphasize the importance of reasoning in 3D for object recognition and detection in complex 3D scenes. In our architecture, this reasoning component is performed by a sub-network which we term the topdown network. This is a simple convolutional network with ResNet-style skip connections which operates on the the 2D feature maps $\orthomap$ generated by the previously described OFT stage. Since the filters of the topdown network are applied convolutionally, all processing is invariant to the location of the feature on the ground plane. This means that feature maps which are distant from the camera receive exactly the same treatment as those that are close, despite corresponding a much smaller region of the image. The ambition is that the final feature representation will therefore capture information purely about the underlying 3D structure of the scene and not its 2D projection.


\subsection{Confidence map prediction}
\label{sec:conf}
Among both 2D and 3D approaches, detection is conventionally treated as a classification problem, with a cross entropy loss used to identify regions of the image which contain objects. In our application however we found it to be more effective to adopt the confidence map regression approach of Huang~\etal~\cite{huang2015densebox}. The confidence map $\confmap(x, z)$ is a smooth function which indicates the probability that there exists an object with a bounding box centred on location $(x, y_0, z)$, where $y_0$ is the distance of the ground plane below the camera. Given a set of $N$ ground truth objects with bounding box centres $\boldsymbol{p_i}=\begin{bmatrix}x_i & y_i & z_i\end{bmatrix}\transpose, i = 1, \dots, N$, we compute the ground truth confidence map as a smooth Gaussian region of width $\sigma$ around the center of each object. The confidence at location $(x, z)$ is given by 

\begin{equation}
\confmap(x, z)=\max_i\exp\left(-\frac{(x_i-x)^2 + (z_i-z)^2}{2\sigma^2}\right).
\end{equation}
The confidence map prediction head of our network is trained via an $\ell_1$ loss to regress to the ground truth confidence for each location on the orthographic grid $\mathcal{H}$. A well-documented challenge is that there are vastly fewer positive (high confidence) locations than negative ones, which leads to the negative component of the loss dominating optimization~\cite{shrivastava2016training, lin2018focal}. To overcome this we scale the loss corresponding to negative locations (which we define as those with $\confmap(x, z) < 0.05$) by a constant factor of $10^{-2}$.

\subsection{Localization and bounding box estimation}

The confidence map $\confmap$ encodes a coarse approximation of the location of each object as a peak in the confidence score, which gives a position estimate accurate up to the resolution $\res$ of the feature maps. In order to localize each object more precisely, we append an additional network output head which predicts the relative offset $\posoffset$ from grid cell locations on the ground plane $(x, y_0, z)$ to the center of the corresponding ground truth object $\boldsymbol{p}_i$:
\begin{equation}
    \posoffset(x, z) = \begin{bmatrix}
        \frac{x_i-x}{\sigma} & \frac{y_i-y_0}{\sigma} & \frac{z_i-z}{\sigma}
    \end{bmatrix}\transpose
    \label{eqn:pos}
\end{equation}
We use the same scale factor $\sigma$ as described in \secref{conf} to normalize the position offsets within a sensible range. A ground truth object instance $i$ is assigned to a grid location $(x, z)$ if any part of the object's bounding box intersects the given grid cell. Cells which do not intersect any ground truth objects are ignored during training. 

In addition to localizing each object, we must also determine the size and orientation of each bounding box. We therefore introduce two further network outputs. The first, the dimension head, predicts the logarithmic scale offset $\dimoffset$ between the assigned ground truth object $i$ with dimensions $\boldsymbol{d}_i=\begin{bmatrix}w_i & h_i & l_i\end{bmatrix}$ and the mean dimensions $\bar{\boldsymbol{d}}=\begin{bmatrix}\bar{w} & \bar{h} & \bar{l}\end{bmatrix}$ over all objects of the given class. 
\begin{equation}
    \dimoffset(x, z) = \begin{bmatrix}
        \log\frac{w_i}{\bar{w}} & \log\frac{h_i}{\bar{h}} & \log\frac{l_i}{\bar{l}}
    \end{bmatrix}\transpose
    \label{eqn:dim}
\end{equation}

The second, the orientation head, predicts the sine and cosine of the objects orientation $\theta_i$ about the y-axis:
\begin{equation}
\angoffset(x, z) = \begin{bmatrix}\sin\theta_i & \cos\theta_i \end{bmatrix}^\top
\label{eqn:ang}
\end{equation}
Note that since we are operating in the orthographic birds-eye-view space, we are able to predict the y-axis orientation $\theta$ directly, unlike other works e.g.~\cite{mousavian20173d} which predict the so-called \emph{observation} angle $\alpha$ to take into account the effects of perspective and relative viewpoint. The position offset $\posoffset$, dimension offset $\dimoffset$ and orientation vector $\angoffset$ are trained using an $\ell_1$ loss.

\begin{table*}[th]
\centering
\caption{Average precision for birds-eye-view ($AP_{BEV}$) and 3D bounding box ($AP_{3D}$) detection on the KITTI test benchmark.}
\label{tab:test}
\begin{tabular}{@{}ccccccccc@{}}
\toprule
\multirow{2}{*}{Method} 		    & \multirow{2}{*}{Modality}     & \multicolumn{3}{c}{$AP_{3D}$}                 & & \multicolumn{3}{c}{$AP_{BEV}$}                  \\ \cmidrule(l){3-5} \cmidrule(l){7-9} 
                       			    &                               & Easy          & Moderate      & Hard          & & Easy            & Moderate      & Hard          \\ \midrule
3D-SSMFCNN~\cite{novak2017vehicle}   & Mono                          & 2.28          & 2.39          & 1.52          & & 3.66            & 3.19          & 3.45          \\
\textbf{OFT-Net (Ours)}            & Mono                          &\textbf{2.50}& \textbf{3.28} & \textbf{2.27}   & & \textbf{9.50}   & \textbf{7.99} & \textbf{7.51} \\ \bottomrule
\end{tabular}

\end{table*}

\begin{table*}[th]
\centering
\caption{Average precision for birds-eye-view ($AP_{BEV}$) and 3D bounding box ($AP_{3D}$) detection on the KITTI validation set.}
\label{tab:val}
\begin{tabular}{@{}ccccccccc@{}}
\toprule
\multirow{2}{*}{Method} 		& \multirow{2}{*}{Modality} & \multicolumn{3}{c}{$AP_{3D}$}                 & & \multicolumn{3}{c}{$AP_{BEV}$}                  \\ \cmidrule(l){3-5} \cmidrule(l){7-9} 
                       			&                           & Easy          & Moderate      & Hard          & & Easy           & Moderate       & Hard          \\ \midrule
3DOP~\cite{chen20153d}  		& Stereo                    & 6.55          & 5.07          & 4.10          & & 12.63          & 9.49           & 7.59          \\ \midrule
Mono3D~\cite{chen2016monocular} & Mono                      & 2.53          & 2.31          & 2.31          & & 5.22           & 5.19           & 4.13          \\
\textbf{OFT-Net (Ours)}                 & Mono                      & \textbf{4.07} & \textbf{3.27} & \textbf{3.29} & & \textbf{11.06} & \textbf{8.79}  & \textbf{8.91} \\ \bottomrule
\end{tabular}

\end{table*}

\subsection{Non-maximum suppression}
Similarly to other object detection algorithms, we apply a non-maximum suppression (NMS) stage to obtain a final discrete set of object predictions. In a conventional object detection setting this step can be expensive since it requires $\mathcal{O}(N^2)$ bounding box overlap computations. This is compounded by the fact that pairs of 3D boxes are not necessarily axis aligned, which makes the overlap computation more difficult compared to the 2D case. Fortunately, an additional benefit of the use of confidence maps in place of anchor box classification is that we can apply NMS in the more conventional image processing sense, i.e. searching for local maxima on the 2D confidence maps $\confmap$. Here, the orthographic birds-eye-view again proves invaluable: the fact that two objects cannot occupy the same volume in the 3D world means that peaks on the confidence maps are naturally separated. 

To alleviate the effects of noise in the predictions, we first smooth the confidence maps by applying a Gaussian kernel with width $\sigma_{NMS}$. A location $(x_i, z_i)$ on the smoothed confidence map $\hat{\confmap}$ is deemed to be a maximum if 
\begin{equation}
    \hat{\confmap}(x_i, z_i) \geq \hat{\confmap}(x_i + m, z_i + n)\quad \forall m, n\in\{-1, 0, 1\}.
\end{equation}
Of the produced peak locations, any with a confidence $S(x_i, y_i)$ smaller than a given threshold $t$ are eliminated. This results in the final set of predicted object instances, whose bounding box center $\boldsymbol{p}_i$, dimensions $\boldsymbol{d}_i$, and orientation $\theta_i$, are given by inverting the relationships in Equations~\ref{eqn:pos}, \ref{eqn:dim} and~\ref{eqn:ang} respectively.


\section{Experiments}

\subsection{Experimental setup}

\paragraph{Architecture} For our front-end feature extractor we make use of a ResNet-18 network without bottleneck layers. We intentionally choose the front-end network to be relatively shallow, since we wish to put as much emphasis as possible on the 3D reasoning component of the model. We extract features immediately before the final three downsampling layers, resulting in a set of feature maps $\{\imagemap^s\}$ at scales $s$ of 1/8, 1/16 and 1/32 of the original input resolution. Convolutional layers with 1$\times$1 kernels are used to map these feature maps to a common feature size of 256, before processing them via the orthographic feature transform to yield orthographic feature maps $\{\orthomap^s\}$. We use a voxel grid with dimensions 80m$\times$4m$\times$80m, which is sufficient to include all annotated instances in KITTI, and set the grid resolution $\res$ to be 0.5m. For the topdown network, we use a simple 16-layer ResNet without any downsampling or bottleneck units. The output heads each consist of a single 1$\times$1 convolution layer. Throughout the model we replace all batch normalization~\cite{ioffe2015batch} layers with group normalization~\cite{wu2018group} which has been found to perform better for training with small batch sizes.

\paragraph{Dataset} We train and evaluate our method using the KITTI 3D object detection benchmark dataset~\cite{geiger2012kitti}. For all experiments we follow the train-val split of Chen~\etal~\cite{chen2016monocular} which divides the KITTI training set into 3712 training images and 3769 validation images.

\paragraph{Data augmentation} Since our method relies on a fixed mapping from the image plane to the ground plane, we found that extensive data augmentation was essential for the network to learn robustly. We adopt three types of widely-used augmentations: random cropping, scaling and horizontal flipping, adjusting the camera calibration parameters $f$ and $(c_u, c_v)$ accordingly to reflect these perturbations.

\paragraph{Training procedure} The model is trained using SGD for 600 epochs with a batch size of 8, momentum of 0.9 and learning rate of $10^{-7}$. Following~\cite{masters2018revisiting}, losses are summed rather than averaged, which avoids biasing the gradients towards examples with few object instances. The loss functions from the various output heads are combined using a simple equal weighting strategy.

\subsection{Comparison to state-of-the-art}
\label{sec:sota}

We evaluate our approach on two tasks from the KITTI 3D object detection benchmark. The 3D bounding box detection task requires that each predicted 3D bounding box should intersect a corresponding ground truth box by at least 70\% in the case of cars and 50\% for pedestrians and cyclists. The birds-eye-view detection task meanwhile is slightly more lenient, requiring the same amount of overlap between a 2D birds-eye-view projection of the predicted and ground truth bounding boxes on the ground plane. At the time of writing, the KITTI benchmark included only one published approach operating on monocular RGB images alone~(\cite{novak2017vehicle}), which we compare our method against in \tabref{test}. We therefore perform additional evaluation on the KITTI validation split set out by Chen~\etal (2016)~\cite{chen2016monocular}; the results of which are presented in \tabref{val}. For monocular methods, performance on the pedestrian and cyclist classes is typically insufficient to obtain meaningful results and we therefore follow other works~\cite{chen2016monocular,chen20153d,novak2017vehicle} and focus our evaluation on the car class only.

It can be seen from Tables~\ref{tab:test} and~\ref{tab:val} that our method is able to outperform all comparable (i.e. monocular only) methods by a considerable margin across both tasks and all difficulty criteria. The improvement is particularly marked on the hard evaluation category, which includes instances which are heavily occluded, truncated or far from the camera. We also show in \tabref{val} that our method performs competitively with the stereo approach of Chen~\etal (2015)~\cite{chen20153d}, achieving close to or in one case better performance than their 3DOP system. This is in spite of the fact that unlike \cite{chen20153d}, our method does not have access to any explicit knowledge of the depth of the scene.  


\subsection{Qualitative results}

\paragraph{Comparison to Mono3D}
We provide a qualitative comparison of predictions generated by our approach and Mono3D~\cite{chen2016monocular} in \figref{mono3d}. A notable observation is that our system is able to reliably detect objects at a considerable distance from the camera. This is a common failure case among both 2D and 3D object detectors, and indeed many of the cases which are correctly identified by our system are overlooked by Mono3D. We argue that this ability to recognise objects at distance is a major strength of our system, and we explore this capacity further in \secref{depth}. Further qualitative results are included in supplementary material.

\paragraph{Ground plane confidence maps}
A unique feature of our approach is that we operate largely in the orthographic birds-eye-view feature space. To illustrate this, \figref{heatmaps} shows examples of predicted confidence maps $\confmap(x, z)$ both in the topdown view and projected into the image on the ground plane. It can be seen that the predicted confidence maps are well localized around each object center.


\afterpage{
\begin{figure*}[p]
    \centering
    \includegraphics[width=\linewidth]{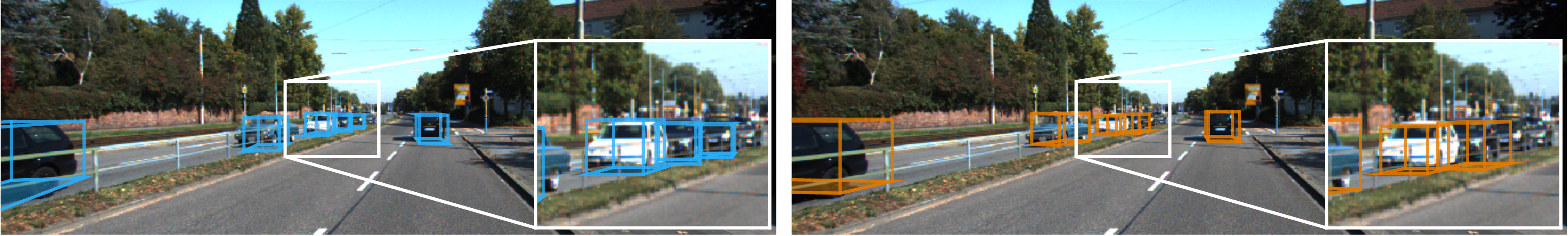}
    \includegraphics[width=\linewidth]{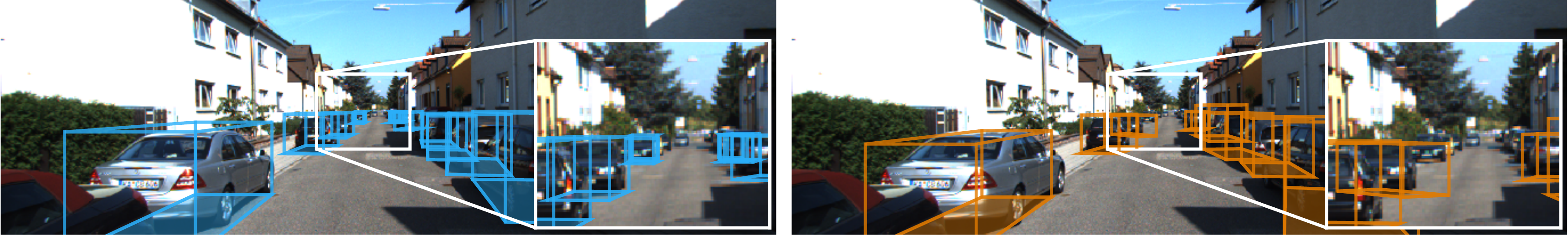}
    \includegraphics[width=\linewidth]{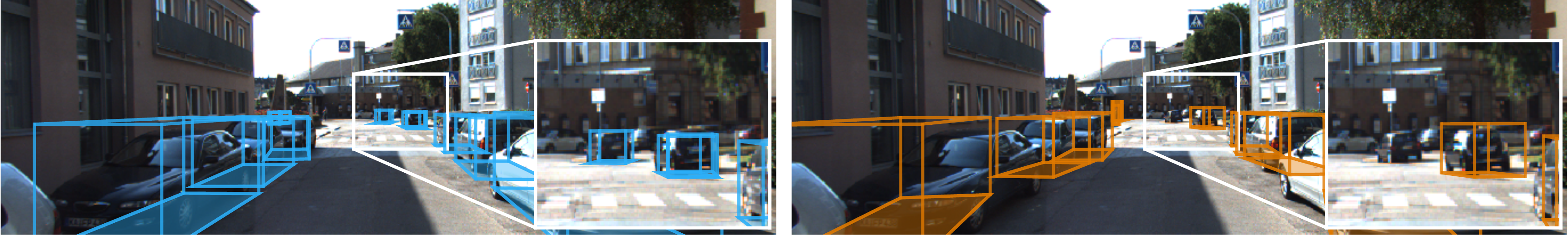}
    \caption{Qualitative comparison between our method (left) and Mono3D \cite{chen2016monocular} (right) on the KITTI validation set. Inset regions highlight the behaviours of the two systems at large distances. We are able to consistently detect distant objects which are beyond the range of Mono3D.} 
    \label{fig:mono3d}
\end{figure*}

\begin{figure*}[p]
    \centering
    \includegraphics[width=.9\linewidth]{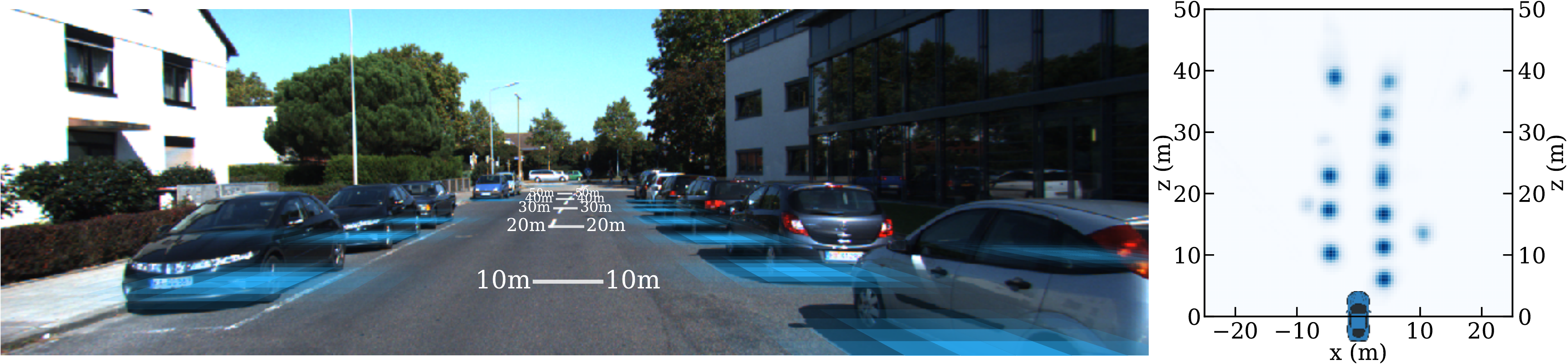} \\
    \vspace{-0.1cm}
    \includegraphics[width=.9\linewidth]{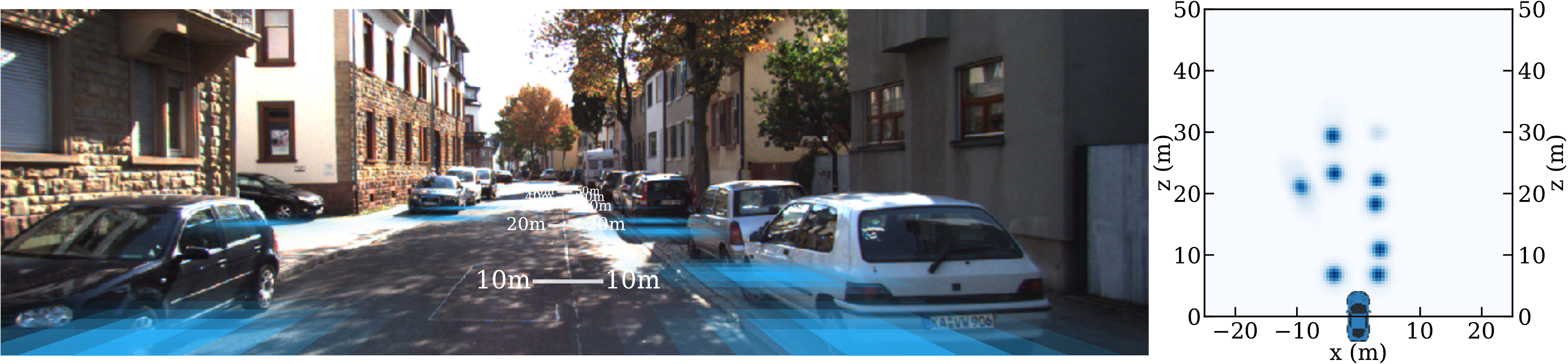} \\
    
    
    \caption{Examples of confidence maps generated by our approach, which we visualize both in birds-eye-view (right) and projected onto the ground plane in the image view (left). We use the pre-computed ground planes of~\cite{chen20153d} to obtain the road position: note that this is for visualization purposes only and the ground planes are not used elsewehere in our approach. Best viewed in color.} 
    \label{fig:heatmaps}
\end{figure*}

\begin{figure*}[p]
    \centering
    \includegraphics[width=1\linewidth]{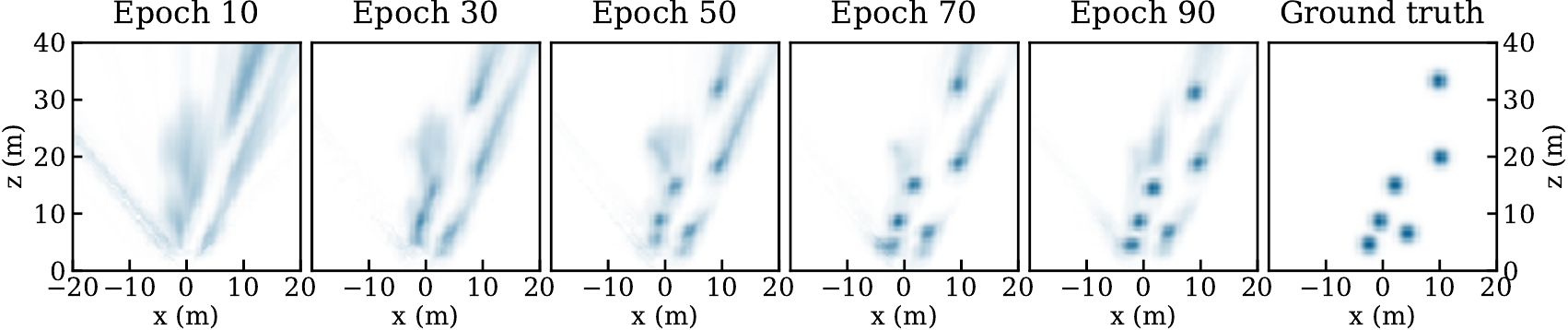}
    \caption{Evolution of ground plane confidence maps during training. The network initially exhibits high uncertainty in the depth direction but gradually resolves this uncertainty as training progresses.}
    \label{fig:evolution}
\end{figure*}

\clearpage
}

\subsection{Ablation study}
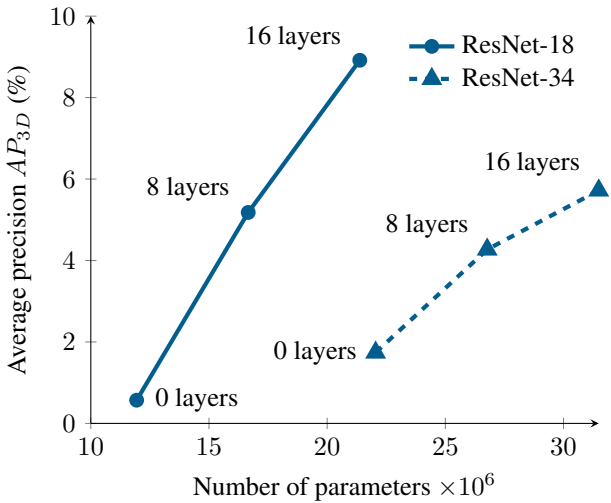
\begin{figure}[ht]
  \centering
  \begin{tikzpicture}
\begin{axis}[
	xlabel=Number of parameters $\times10^6$,
    ylabel=Average precision $AP_{3D}$ (\%),
    ymin=0, ymax=10, xmin=10,
    height=7cm,
    width=\linewidth
	]
    \addplot[blue, mark=*] table [x=Parameters,y=Resnet18, col sep=comma]{data/ablation.txt};
    \addlegendentry{ResNet-18}
    \addplot[blue, mark=triangle*, mark size=3pt, mark options=solid, dashed] table [x=Parameters,y=Resnet34, col sep=comma]{data/ablation.txt};
    \addlegendentry{ResNet-34}
    
    \node[label={0:{0 layers}}] at (axis cs:11.94, 0.57) {};
    \node[label={165:{8 layers}}] at (axis cs:16.66, 5.18) {};
    \node[label={170:{16 layers}}] at (axis cs:21.38, 8.917) {};
    
    \node[label={180:{0 layers}}] at (axis cs:22.05, 1.74) {};
    \node[label={165:{8 layers}}] at (axis cs:26.77, 4.27) {};
    \node[label={150:{16 layers}}] at (axis cs:31.49, 5.72) {};
\end{axis}
\end{tikzpicture}
  \caption{Ablation study showing the effect of reducing the number of layers in the topdown network on performance for two different frontend architectures. Zero layers implies that topdown network has been removed entirely.}
  \label{fig:ablation}
  \vspace{-0.4cm}
\end{figure}

A central claim of our approach is that reasoning in the orthographic birds-eye-view space significantly improves performance. To validate this claim, we perform an ablation study where we progressively remove layers from the topdown network. In the extreme case, when the depth of the topdown network is zero, the architecture is effectively reduced to RoI pooling~\cite{girshick2015fast} over projected bounding boxes, rendering it similar to R-CNN-based architectures. Figure~\ref{fig:ablation} shows a plot of average precision against the total number of parameters for two different architectures. 

The trend is clear: removing layers from the topdown network significantly reduces performance. Some of this decline in performance may be explained by the fact that reducing the size of the topdown network reduces the overall depth of the network, and therefore its representational power. However, as can be seen from Figure~\ref{fig:ablation}, adopting a shallow front-end (ResNet-18) with a large topdown network achieves significantly better performance than a deeper network (ResNet-34) without any topdown layers, despite the two architectures having roughly the same number of parameters. This strongly suggests that a significant part of the success of our architecture comes from its ability to reason in 3D, as afforded by the 2D convolution layers operating on the orthographic feature maps.   

\section{Discussion}
\subsection{Performance as a function of depth}
\label{sec:depth}
\begin{figure}[h!]
  \centering
  \begin{tikzpicture}
\begin{axis}[
	xlabel=Minimum depth (m),
    ylabel=Average precision $AP_{BEV}$ (\%),
    ymax=45,
    height=6cm,
    width=\linewidth
	]
    \addplot[smooth,orange] table [x=Depth,y=Mono3D]{data/ap-vs-depth.txt};
    \addlegendentry{Mono3D~\cite{chen2016monocular}}
    \addplot[smooth,blue] table [x=Depth,y=Ours]{data/ap-vs-depth.txt};
    \addlegendentry{OFT-Net (Ours)}
\end{axis}
\end{tikzpicture}
\vspace{-1em}
  \caption{Average BEV precision (val) as a function of the minimum distance of objects from the camera. We use an IoU threshold of 0.5 to better compare performance at large depths.}
  \label{fig:depth}
\end{figure}
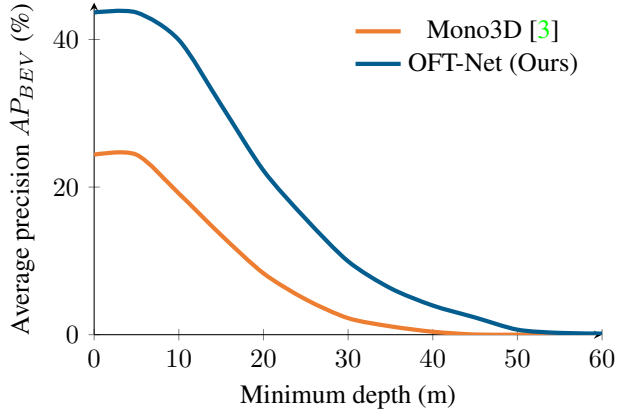

Motivated by the qualitative results in Section~\ref{sec:sota}, we wished to further quantify the ability of our system to detect and localize distant objects. Figure~\ref{fig:depth} plots performance of each system when evaluated only on objects which are at least the given distance away from the camera. Whilst we outperform Mono3D over all depths, it is also apparent that the performance of our system degrades much more slowly as we consider objects further from the camera. We believe that this is a key strength of our approach.

\subsection{Evolution of confidence maps during training}

While the confidence maps predicted by our network are not necessarily calibrated estimates of model certainty, observing their evolution over the course of training does give valuable insights into the learned representation. \figref{evolution} shows an example of a confidence map predicted by the network at various points during training. During the early stages of training, the network very quickly learns to identify regions of the image which contain objects, which can be seen by the fact that high confidence regions correspond to projection lines from the optical center at $(0, 0)$ which intersect a ground truth object. However, there exists significant uncertainty about the depth of each object, leading to the predicted confidences being blurred out in the depth direction. This fits well with our intuition that for a monocular system depth estimation is significantly more challenging than recognition. As training progresses, the network is increasingly able to resolve the depth of the objects, producing sharper confidence regions clustered about the ground truth centers. It can be observed that even in the latter stages of training, there is considerably greater uncertainty in the depth of distant objects than that of nearby ones, evoking the well-known result from stereo that depth estimation error increases quadratically with distance.   



\section{Conclusions}
In this work we have presented a novel approach to monocular 3D object detection, based on the intuition that operating in the birds-eye-view domain alleviates many undesirable properties of images which make it difficult to infer the 3D configuration of the world. We have proposed a simple \emph{orthographic feature transform} which transforms image-based features into this birds-eye-view representation, and described how to implement it efficiently using integral images. This was then incorporated into part of a deep learning pipeline, in which we particularly emphasized the importance of spatial reasoning in the form of a deep 2D convolutional network applied to the extracted birds-eye-view features. Finally, we experimentally validated our hypothesis that reasoning in the topdown space does achieve significantly better results, and demonstrated state-of-the-art performance on the KITTI 3D object benchmark.

\FloatBarrier
\bibliographystyle{ieee}
\bibliography{references}

\end{document}